%% file: main.tex
\documentclass[a4paper]{coupled2021}

\usepackage[hyphens]{url}  
\usepackage{graphicx} 
\usepackage{caption} 

\usepackage{comment}
\excludecomment{omitext}

\usepackage{times,colordvi,bbm}
\usepackage{wrapfig}
\usepackage{tikz}
\usetikzlibrary{patterns} 
\usepackage{subfig}
\usepackage[latin1]{inputenc}
\usepackage[fleqn]{amsmath}
\usepackage{amsfonts,amssymb,amsbsy,amscd,amsthm}
\usepackage[mathscr]{eucal}
\usepackage{booktabs}
\usepackage{tabularx,multirow}
    \newcolumntype{L}{>{\raggedright\arraybackslash}X}
    \newcolumntype{C}{>{\centering\arraybackslash}X}
\usepackage{mathrsfs}
\usepackage{array}
\usepackage{verbatim}
\usepackage{enumitem}
\usepackage[linesnumbered,ruled,lined,boxed]{algorithm2e}
\RestyleAlgo{boxruled}
\SetKwRepeat{Do}{do}{while} 

\usepackage{listings}
\usepackage[off]{auto-pst-pdf}
\usepackage{ifthen}
\usepackage{mathtools}
\DeclarePairedDelimiter{\nm}{\lVert}{\rVert}

\mathchardef\breakingcomma\mathcode`\,
{\catcode`,=\active
  \gdef,{\breakingcomma\discretionary{}{}{}}
}

\DeclareMathOperator*{\argmin}{argmin} 

\renewcommand{\vec}[1]{{\ensuremath{\boldsymbol{\mathrm #1}}}}
\newcommand{\vdot}{{\ifnum\thedotStyle=0{\ensuremath\cdot}\else
    {\boldsymbol{\mathsf{\ensuremath\cdot}}}\fi}}

\newcommand{\bz}{\mathbf{z}}

\newcommand{\bo}{\boldsymbol{\omega}}

\newcommand{\pd}[2]{\frac{\partial{#1}}{\partial{#2}}}

\newcommand{\ddt}[1]{\frac{d \,{#1}}{dt}}

\newcommand{\R}{\ensuremath{\mathbb{R}}}

\newcommand{\strongRes}{\boldsymbol{\mathfrak{R}}}

\theoremstyle{remark}

\title{DATA-DRIVEN REDUCED ORDER MODELING OF ENVIRONMENTAL HYDRODYNAMICS USING DEEP AUTOENCODERS AND NEURAL ODES}

\author{SOURAV DUTTA$^{1\dagger}$, PETER RIVERA-CASILLAS$^{1}$, ORIE M. CECIL$^{1}$, MATTHEW W. FARTHING$^{1}$, EMMA PERRACCHIONE$^{2}$, AND MARIO PUTTI$^{3}$}

\heading{S. Dutta, P. Rivera-Casillas, O. M. Cecil, M. W. Farthing, E. Perracchione, and M. Putti}

\address{$^{1}$ USACE Engineer Research Development Center, 3909 Halls Ferry Rd, Vicksburg MS 39180, USA,
\and
$^{2}$ University of Genoa, Via Dodecaneso 35, 16146 Genova, Italy,
\and
$^{3}$ University of Padua, Via Trieste, 63, 35131 Padova, Italy,
\and
$^{\dagger}$ email: sourav.dutta@erdc.dren.mil }


\abstract{Model reduction for fluid flow simulation continues to be of great interest across a 
number of scientific and engineering fields. In a previous work \cite{Dutta2021-AAAI}, 
we explored the use of Neural Ordinary Differential Equations (NODE) as a non-intrusive method 
for propagating the latent-space dynamics in reduced order models.
Here, we investigate employing deep autoencoders for discovering the reduced basis 
representation, the dynamics of which are then approximated by NODE. The ability of deep 
autoencoders to represent the latent-space is compared to the traditional proper orthogonal 
decomposition (POD) approach, again in conjunction with NODE for capturing the dynamics. Additionally, we compare their behavior with two classical non-intrusive methods based on POD and radial basis function interpolation as well as dynamic mode decomposition.
The test problems we consider include incompressible flow around a cylinder as well as a real-world 
application of shallow water hydrodynamics in an estuarine system. Our findings 
indicate that deep autoencoders can leverage nonlinear manifold learning to achieve a highly efficient compression of spatial information and 
define a latent-space that appears to be more suitable for capturing the temporal dynamics through 
the NODE framework.
}

\begin{document}


\section{INTRODUCTION}
\input{alt_Intro}

\section{METHODOLOGY}
The standard ROM development framework can be divided into three stages:
\begin{enumerate}
    \item identification of a low-dimensional latent (or reduced-order) space,
    \item representation of the nonlinear dynamical system in terms of the reduced basis and modeling the evolution of the system of modal coefficients, and
    \item reconstruction in the high-fidelity space for validation and analysis.
\end{enumerate}

\subsection{Dimension reduction}
In this work, the dominant features of the unsteady flow-field have been extracted using a linear modal decomposition technique, POD, and a nonlinear manifold learning method that relies on deep, fully-connected, multi-layer perceptron (MLP) autoencoders that are highly expressive and scalable \cite{HR2006}. 
POD is a popular technique for dimension reduction of the solution manifold of a dynamical system by determining a linear reduced space spanned by an orthogonal basis with an associated energetic hierarchy, and which represents an optimal approximation of the solution manifold with respect to the $L^2$-norm. Given a matrix of high-fidelity system snapshots $\vec S \in \R^{N\times M}$ and a matrix of orthonormal POD basis vectors $\vec \theta$, the modal coefficient matrix $\vec Z = \vec \Theta^T \vec S $ constitutes our training data for the latent space learning methods. \cite{Taira2020} provides an excellent overview of POD as well as a comparison with other dimension-reduction techniques.

\subsubsection{Autoencoders}
An autoencoder is a type of feedforward neural network that is designed to learn the identity mapping, $\vec h : \vec v \mapsto \widetilde{\vec v}$ such that $\widetilde{\vec v} \approx \vec v$ and $\vec h: \mathbb{R}^N \mapsto \mathbb{R}^N$. This is accomplished using a two-part architecture. The first part is called an encoder, $\vec h_E$, defined by $\vec z = \vec h_E(\vec v; \vec \theta_E )$ where $\vec z \in \mathbb{R}^m$ ($m \ll N$), which maps a high-dimensional input vector $\vec v$ to a low-dimensional latent vector $\vec z$. 
\begin{wrapfigure}{r}{0.5\textwidth}
    \centering
    \includegraphics[width=0.5\columnwidth]{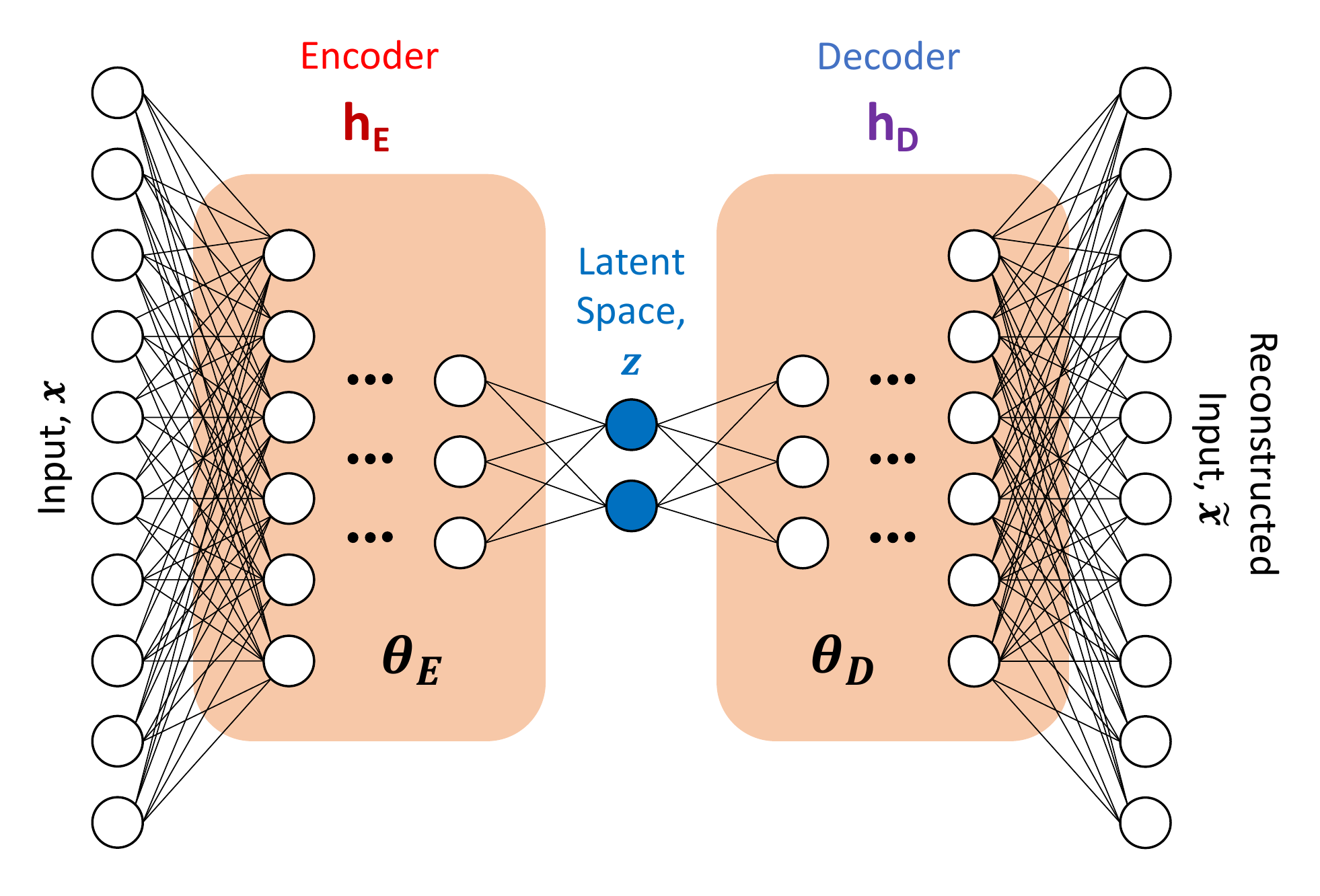}
    \caption{The architecture of the autoencoder network}
    \label{fig:ae_network}
\end{wrapfigure}
The second part is called a decoder, $\vec h_D$, defined as $\widetilde{\vec v} = \vec h_D (\vec z; \vec \theta_D)$, which maps the latent vector $\vec z$ to an approximation $\widetilde{\vec v}$ of the high-dimensional input vector $\vec v$. The combination of the two parts yields an autoencoder of the form 
\begin{align}
  \vec h: \vec v \mapsto \vec h_D \circ \vec h_E(\vec v).  
\end{align}
This autoencoder is trained by computing the optimal values of the parameters $(\vec \theta_E^*, \vec \theta_D^*)$ that minimize the reconstruction error over all the training data
\begin{align}
    \vec \theta_E^*, \, \vec \theta_D^* = \argmin_{\vec \theta_E, \vec \theta_D} \mathcal{L}(\vec v, \widetilde{\vec v}),
\end{align}
where $\mathcal{L}(\vec v, \widetilde{\vec v})$ is a chosen measure of discrepancy between $\vec v$ and its approximation $\widetilde{\vec v}$.
The restriction $\dim({\vec z}) = m \ll N = \dim({\vec v})$ forces the autoencoder model to learn the salient features of the input data via compression into a low-dimensional space and then reconstructing the input, instead of directly learning the identity function. It is worth noting that with the choice of a linear, single-layer encoder of the form $\vec z = H_E \vec v$, and a linear, single-layer decoder of the form $\widetilde{\vec v} = \vec H_D \vec z$, where $\vec H_E \in \mathbb{R}^{m \times N}$, $\vec H_D \in \mathbb{R}^{N \times m}$, and a squared reconstruction error as the loss function $\mathcal{L}(\widetilde{\vec v}, \vec v) = \nm{\vec v - \widetilde{\vec v}}_2^2$, the autoencoder model has been shown to learn the same subspace as that spanned by the first $m$ POD modes if $\vec H = \vec H_E = \vec H_D$. However, additional constraints are necessary to ensure that the columns of $\vec H$ form an orthonormal basis and follow an energy-based hierarchical ordering \cite{P2018}.

In this work, autoencoders are employed to generate separate low-dimensional latent representations of the pressure (depth) and the velocity snapshot data obtained from the high-fidelity simulation of the numerical examples. The encoder and decoder neural networks are constructed using fully-connected MLP architectures, as depicted in Figure \ref{fig:ae_network}. As the high-fidelity simulation data is usually available on a two-dimensional spatial grid, the data is first flattened and then fed to the autoencoder model.

\subsection{Latent space evolution}

In this section, we outline the non-intrusive framework for modeling the evolution of time-series data in the latent space. The first method employs RBF interpolation which is a classical, data-driven, kernel-based method for computing an approximate continuous response surface that aligns with the given multivariate data. More details about the POD-RBF NIROM framework can be found in \cite{DFPSP2021}. The second technique called NODE is a neural-network based method to predict the continuous evolution of a vector $\vec c$ over time, that is designed to preserve memory effects within the architecture.

\subsubsection{Neural ordinary differential equations}
We assume that the time evolution of the modal coefficients of the high-fidelity dynamical system in the latent space can be modeled using a (first-order) ODE, 
\begin{align}\label{node-ivp}
\ddt{\bz} = \mathcal{F}(t, \bz(t)), \text{ with } \bz(0) = \bz^0, \; \bz \in \R^d, d \geq 1.
\end{align}
The goal is to obtain a NN approximation $\widehat{\mathcal{F}}$ of the dynamics function $\mathcal{F}$ such that $\ddt{\bz} \approx \text{net} (t,\bz) = \widehat{\mathcal{F}}(t, \bz, \bo)$. The full procedure can be outlined as follows:
\begin{enumerate}
    \item Compute the time series of modal coefficients $[\bz^0, \ldots, \bz^{M-1}]$ for $t \in \{0,\ldots,M-1\}$ where $\bz^k \in \R^m$.
    \item Initialize a NN approximation for the dynamics function $\widehat{\mathcal{F}}(t, \bz, \bo)$ where $\bo$ represents the initial NN parameters.
    \item The NN parameters are optimized iteratively through the following steps.
    \begin{enumerate}
        \item Compute the approximate forward time trajectory of the modal coefficients by solving eq. (\ref{node-ivp}) using a standard ODE integrator as,
    \begin{align}
        \hat{\bz}^{M-1} = ODESolve(\widehat{\mathcal{F}}, \bo, \bz^0, t^0, t^{M-1})
    \end{align}
    \item The free parameters of the NODE model are $\{\bo,t^0, t^{M-1}\}$. Evaluate the differentiable loss function 
    $\mathcal{L}\left(ODESolve(\widehat{\mathcal{F}}, \bo, \bz^0, t^0, t^{M-1}) - \bz^{M-1}\right)$.
    
    \item To optimise the loss, compute gradients with respect to the free parameters. Similar to the usual backpropagation algorithm, this can be achieved by first computing the gradient $\partial \mathcal{L}/\partial \widehat{\bz}(t)$, and then a performing a reverse traversal through the intermediate states of the ODE integrator. For a memory-efficient implementation, the adjoint method \cite{Chen2018} can be used to backpropagate the errors by solving an adjoint system for the augmented state vector $\vec b = [\pd{\mathcal{L}}{\widehat{\bz}}, \pd{\mathcal{L}}{\bo}, \pd{\mathcal{L}}{t}]^T $ backwards in time from $t^{M-1}$ to $t^0$. 
    \item The gradient $\pd{\mathcal{L}}{\bo}(t=0)$ computed in the previous step is used to update the parameters $\bo$ by using an optimization algorithm like RMSProp or Adam.
    \end{enumerate}
    \item The trained NODE approximation of the dynamics function can be used to compute predictions for the time trajectory of the modal coefficients. 
\end{enumerate}

Following \cite{Dutta2021-AAAI}, we adopt the TFDiffEq (\url{https://github.com/titu1994/tfdiffeq}) library that runs on the Tensorflow Eager Execution platform to train the NODE models. 
RMSProp is adopted for loss minimization with an initial learning rate of $0.001$, a staircase decay function with a range of variable decay schedules, and a momentum coefficient of $0.9$. 

As a final point of comparison, we consider the standard Dynamic mode decomposition (DMD)\cite{Schmid2010,KBBP2016} algorithm, which is a powerful data-driven method capable of providing an accurate decomposition of a complex system into spatiotemporal coherent structures that may be used for short-time future-state prediction.

\section{NUMERICAL EXPERIMENTS}
In this section, we first assess the use of autoencoders for building a reduced space in which the system dynamics are propagated by NODE for a benchmark flow problem characterized by the incompressible Navier Stokes equations (NSE). We compare the performance of this framework with a POD-NODE method where NODE is employed to capture the evolution of modal coefficients in a reduced space defined by a POD basis. Moreover, we also examine the relative performance of different NIROM models for a real-world estuarine flow application governed by the shallow water equations (SWE). The POD-RBF and DMD NIROM training runs were performed on a Macbook Pro 2018 with a $2.9$ GHz 6-Core Intel Core i9 processor and 32 GB 2400 MHz DDR4 RAM. The autoencoder latent-space representations were trained on Vulcanite, a high performance computer at the U.S. Army Engineer Research and Development Center DoD Supercomputing Resource Center (ERDC-DSRC). Vulcanite is equipped with NVIDIA Tesla V100 PCIe GPU accelerator nodes and has 32GB memory/node. Training for the NODE models was performed in serial on Jim, a high performance computer at the U.S. Army Engineer Research and Development Center Coastal and Hydraulics Lab (CHL), which is equipped with 2 Intel Xeon E5-2699 v3 CPUs and 128Gbytes of memory/node.


\subsection{Flow around a cylinder}
The problem of two-dimensional, incompressible flow around a cylinder is a classical benchmark CFD example that simulates a time-periodic fluid flow through a 2D pipe with a circular obstacle. For further details about the problem setup please see \cite{Dutta2021-AAAI}.
High-fidelity simulation data is obtained with OpenFOAM using an unstructured mesh with $14605$ nodes at $Re=100$, such that the flow exhibits the periodic shedding of von Karman vortices. $313$ training snapshots are collected for $t=[2.5,5.0]$ seconds with $\Delta t=0.008$ seconds, and the NIROM predictions are obtained for $t=[2.5,6.0]$ seconds with $\Delta t=0.002$ seconds.

\begin{table*}[ht]
  \centering
  \begin{tabular}{l m{1.4cm} m{1.4cm} m{1.5cm} c c r}
  \toprule
    Id  & Units & Encoder & Decoder  & Scaling  & MSE     & Training\\
    \toprule
    Range & ($p,v_x,v_y$) 2-8 & linear, relu, ... & elu, tanh, ...  &  &  & \\
    \midrule 
    AE1  & $(5,8,7)$   & linear  & sigmoid   & $[0,1]$         & 2.398e-6 & 9.38 min \\
    AE2  & $(2,2,2)$   & linear  & sigmoid   & $[0,1]$         & 4.196e-6 & 9.56 min \\
    AE3  & $(2,2,2)$   & linear  & tanh      & $[-1,1]$        & 2.499e-6 & 9.29 min \\
    AE4  & $(2,3,3)$   & linear  & sigmoid   & $[0,1]$         & 3.107e-6 & 9.07 min \\
    \bottomrule
  \end{tabular}
  \caption{Best Autoencoder architectures for the cylinder example. Models were trained for $1000$ epochs using the Adam optimizer with initial learning rate = 1e-3 and momentum = $0.9$.}
  \label{tab:ae_cylinder}
\end{table*}

Several different architectures were explored for the autoencoder model by varying the model parameters like the dimension of the latent space, the activation functions, various configurations of the learning rate evolution during training, and Table \ref{tab:ae_cylinder} shows four of the best architectures for the cylinder flow example. The second column shows the size of the latent space for each of the solution variables $p, v_x, v_y$. The third and fourth columns list the activation functions used in the last hidden encoder and the last hidden decoder layers, respectively. 
\begin{wrapfigure}{r}{0.6\textwidth}
  \begin{center}
    \includegraphics[width=0.6\columnwidth]{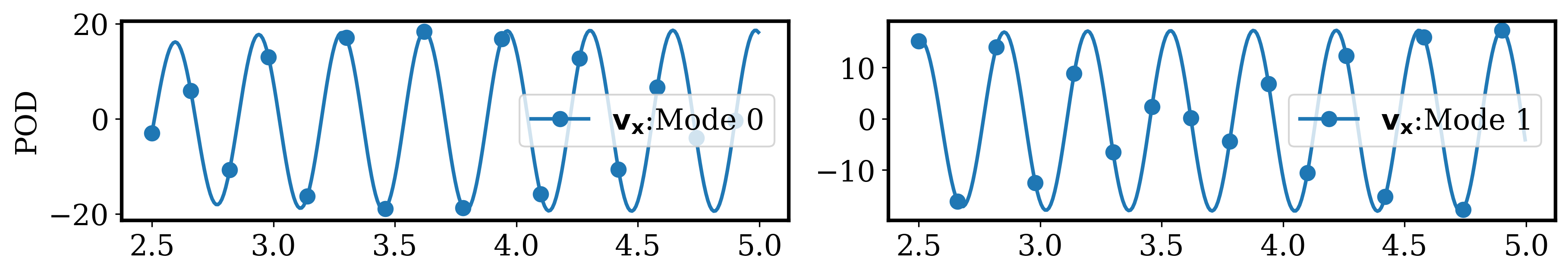}\\
    \includegraphics[width=0.6\columnwidth]{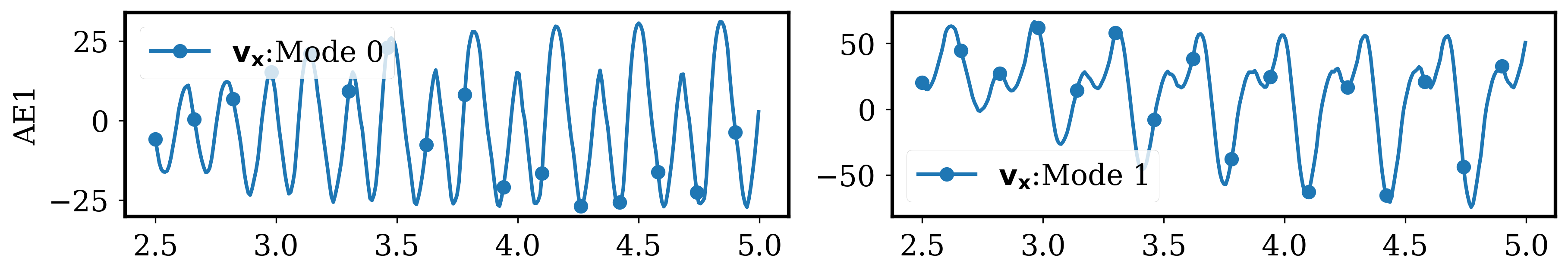}\\
    \includegraphics[width=0.6\columnwidth]{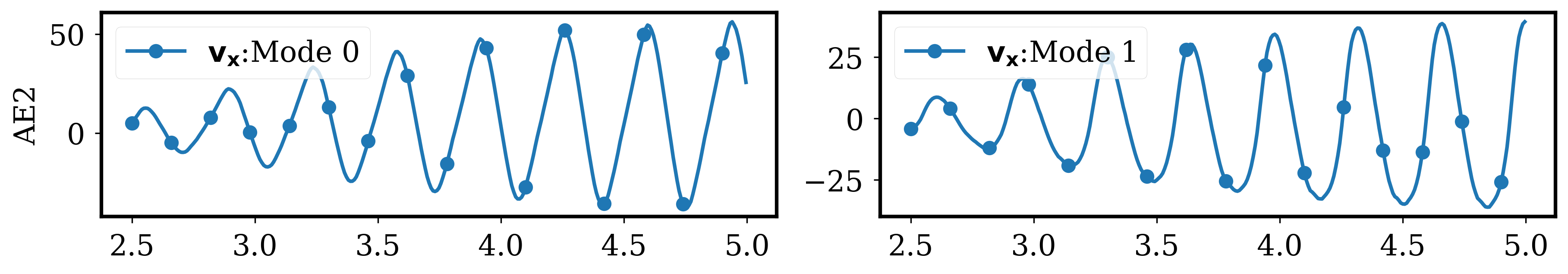}\\
    \includegraphics[width=0.6\columnwidth]{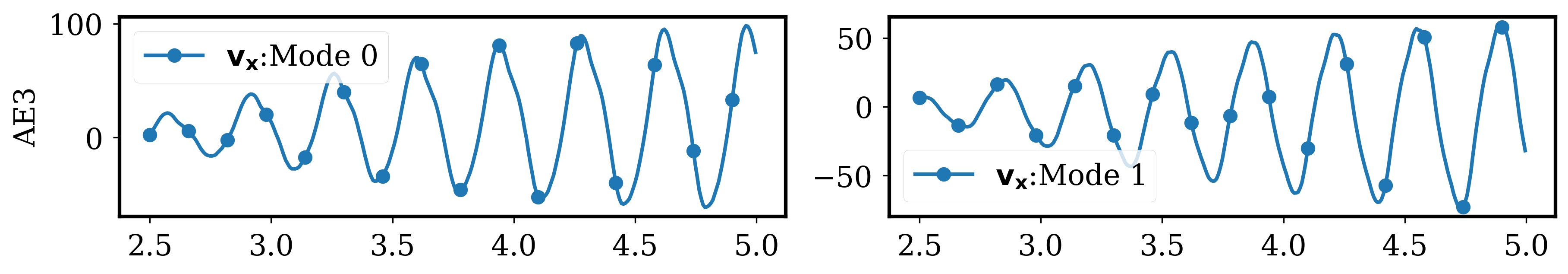}\\
    \includegraphics[width=0.6\columnwidth]{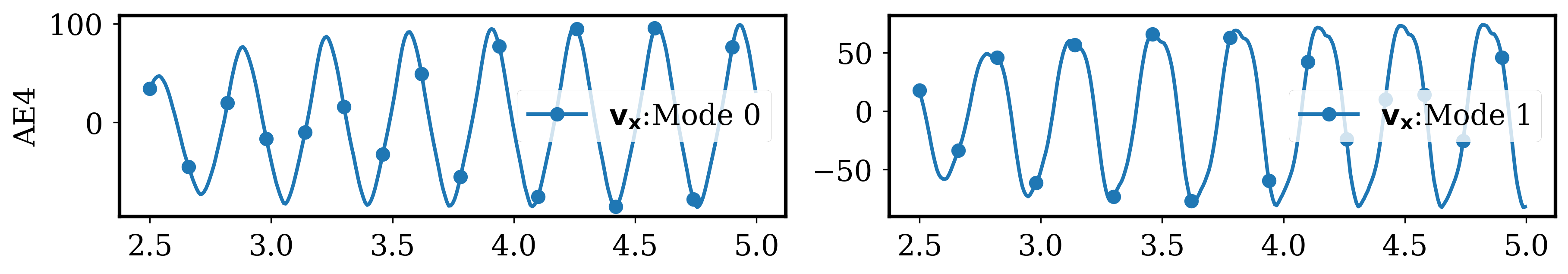}\\
    \end{center}
  \caption{Visualization of the modal coefficients of the first two latent space modes for x-velocity, as generated by POD and the four chosen autoencoder models for the cylinder example}\label{fig:cylinder_aepod_modes}
\end{wrapfigure}
The fifth column shows the scaling applied to the input data which is directly determined by the activation function used in the decoder output layer. In general, it was found that activation functions that required scaled input data like \textit{sigmoid}, \textit{tanh} performed better for the decoder output layer than some of the (semi-)unbounded ones like \textit{linear}, \textit{relu}, and \textit{elu}. However, unbounded activation functions like \textit{linear} and \textit{elu} were seen to generate a more accurate and efficient latent space. The encoder and the decoder networks were made up of four hidden layers with gradually decreasing and gradually increasing size, respectively, and characterized by the \textit{relu} activation function, which were found to generate the least noisy latent representations. The \textit{Adam} optimizer is used for training with an initial learning rate=$10^{-3}$ and momentum=$0.9$. An adaptive learning rate decay algorithm is employed that monitors the training loss and reduces the learning rate by a factor of 2 if no improvement is detected for $200$ epochs. The sixth column of Table \ref{tab:ae_cylinder} lists the total mean square reconstruction error for all three solution variables, while the last column shows the training times for each model on two NVIDIA Tesla V100 GPU nodes. 

\begin{figure}[htb]
    \centering
    \includegraphics[width=0.9\columnwidth]{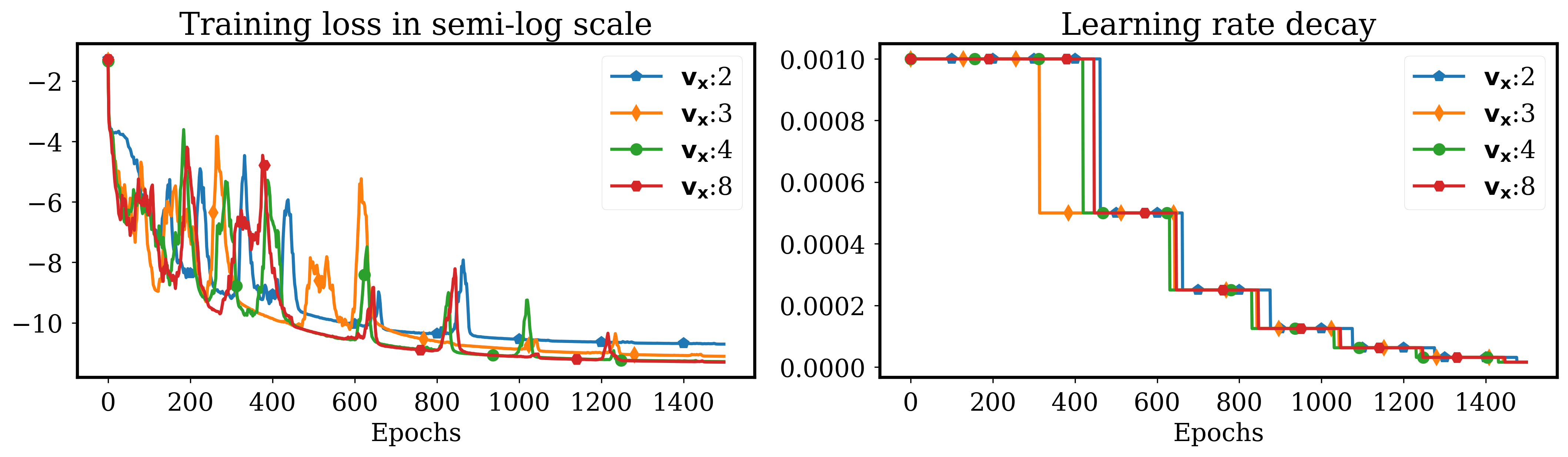}
    \caption{Training characteristics for a fixed autoencoder architecture while varying the latent space dimension}
    \label{fig:ae_training_varying_dim}
\end{figure}

\begin{figure}[htb]
    \centering
    \includegraphics[width=0.9\columnwidth]{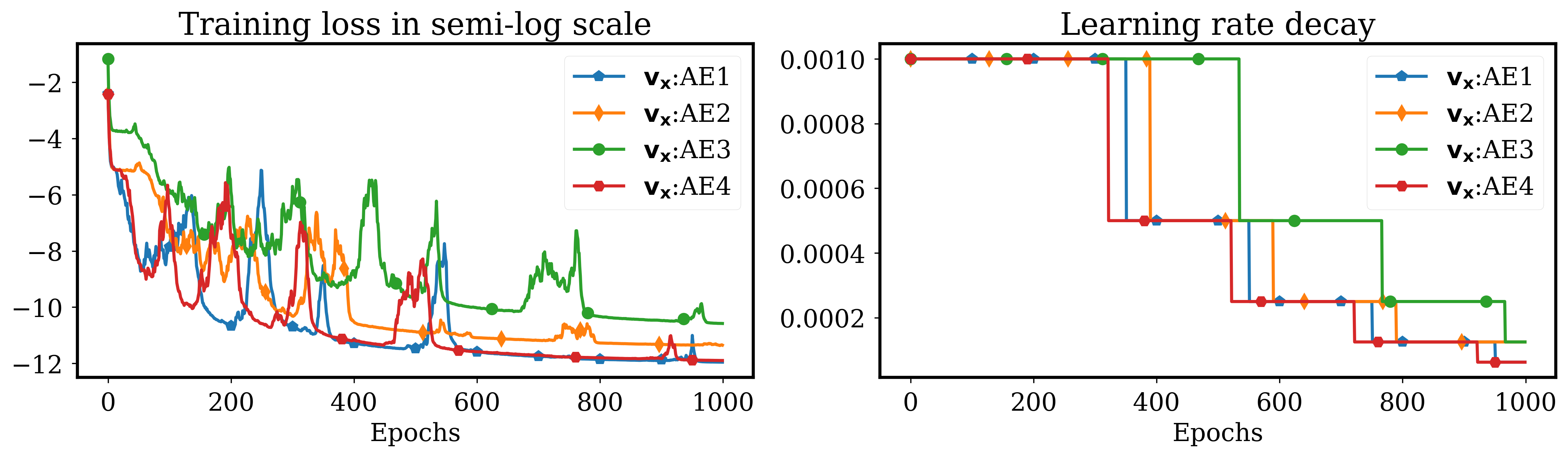}
    \caption{Training characteristics of the chosen four autoencoder architectures (see Table \ref{tab:ae_cylinder}) over 1000 epochs }
    \label{fig:ae_training_varying_models}
\end{figure}

Figure \ref{fig:cylinder_aepod_modes} shows the temporal coefficients for the first two latent modes of $v_x$ using a POD truncation that captures $99\%$ of modal energy content, and the four chosen AE models. While the POD modal coefficients are arranged according to the decreasing order of amplitude, this cannot be guaranteed for the AE-generated spaces. AE2 and AE3 models define a latent space of dimension 2, whereas the AE4 model has 3 latent space units and the AE1 model is identical to the dimension of the POD bases for each variable - 5 ($p$), 8 ($v_x$), and 7($v_y$). The richer quality of the AE1 latent space leads to better expressivity which is reflected in the subtler features captured in the modal coefficients (Figure \ref{fig:cylinder_aepod_modes} row 2) and also in the lowest reconstruction error (Table \ref{tab:ae_cylinder}).

Figure \ref{fig:ae_training_varying_dim} shows the evolution of the training loss and the adaptive decay of the learning rate while training autoencoder models for $v_x$ using four gradually increasing latent space dimensions - $2,3,4,8$. The AE3 architecture was adopted for these runs. The optimization of the hyperparameters with respect to the training loss becomes gradually harder as the dimension of the latent space increases. So the models with smaller latent spaces initially show a faster reduction in training losses. However, the enhanced expressivity of the models with higher latent dimensions allow them to reach lower values of training loss after sufficient epochs of training, whereas the losses for the models with smaller latent dimensions appear to stagnate. Figure \ref{fig:ae_training_varying_models} shows the training loss and the learning rate decay for the four chosen AE models. The AE1 and the AE4 models with the relatively richer latent spaces achieve the sharpest reduction in training loss, whereas the latent space dimensions for models AE2 and AE3 appear to be significant barriers in their training efficiency.

\begin{table*}[t]
  \centering
  \begin{tabular}{l c c m{1.1cm} m{1.79cm} c c c r}
  \toprule
    Id & Lyrs & Units & Act.  & LR steps, rate & Scaling & Aug. & MSE     & Training\\
    \toprule
    Range & 1-4 & 32-512 & tanh, elu,... & 5k-25k, 0.1-0.9 &  & & & \\
    \midrule 
    AE1-NODE1  & 1      & 256   & elu   & 10k, 0.3  & No      & No        & 2.30e-5 & 28.80 hrs \\
    AE1-NODE2  & 1      & 256   & tanh  & 5k,  0.7  & Yes     & No        & 1.34e-4 & 28.69 hrs \\
    AE1-NODE3  & 1      & 512   & elu   & 5k, 0.5   & No      & No        & 1.97e-5 & 29.17 hrs \\
    AE1-NODE4  & 1      & 256   & tanh  & 10k, 0.25 & Yes     & Yes       & 1.49e-4 & 28.27 hrs \\
    AE1-NODE5  & 4      & 64    & tanh  & 5k, 0.5   & Yes     & No        & 1.33e-4 & 33.08 hrs \\
    \bottomrule
  \end{tabular}
  \caption{Results for the best NODE architectures for the cylinder example using the latent space defined by the AE1 autoencoder model. All NODE models were trained for 50000 epochs using the fourth-order Runge-Kutta solver and the RMSProp optimizer with initial learning rate = 1e-3 and momentum = 0.9.}
  \label{tab:node_cylinder}
\end{table*}

An extensive exploration of the NODE hyperparameter and architecture space for the cylinder example was reported in \cite{Dutta2021-AAAI}. Based on those inferences and some further numerical study, five architectures were selected, and the results after training for 50000 epochs using the latent space data generated by the AE1 model are presented in Table \ref{tab:node_cylinder}. 
All the models generate accurate predictions at a finer temporal resolution than the training data, and have excellent agreement with the high-fidelity solution even while extrapolating outside the training data ($5 \leq t \leq 6$ seconds).
\begin{figure}[htb]
  \begin{center}
    \includegraphics[width=0.9\columnwidth]{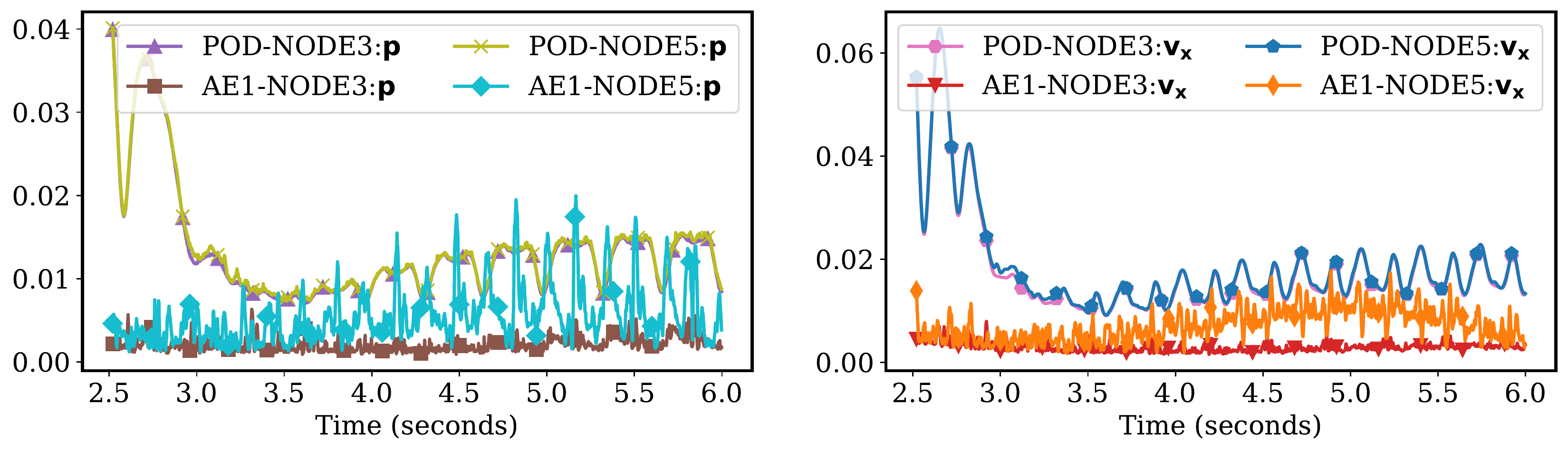}\\
    \includegraphics[width=0.9\columnwidth]{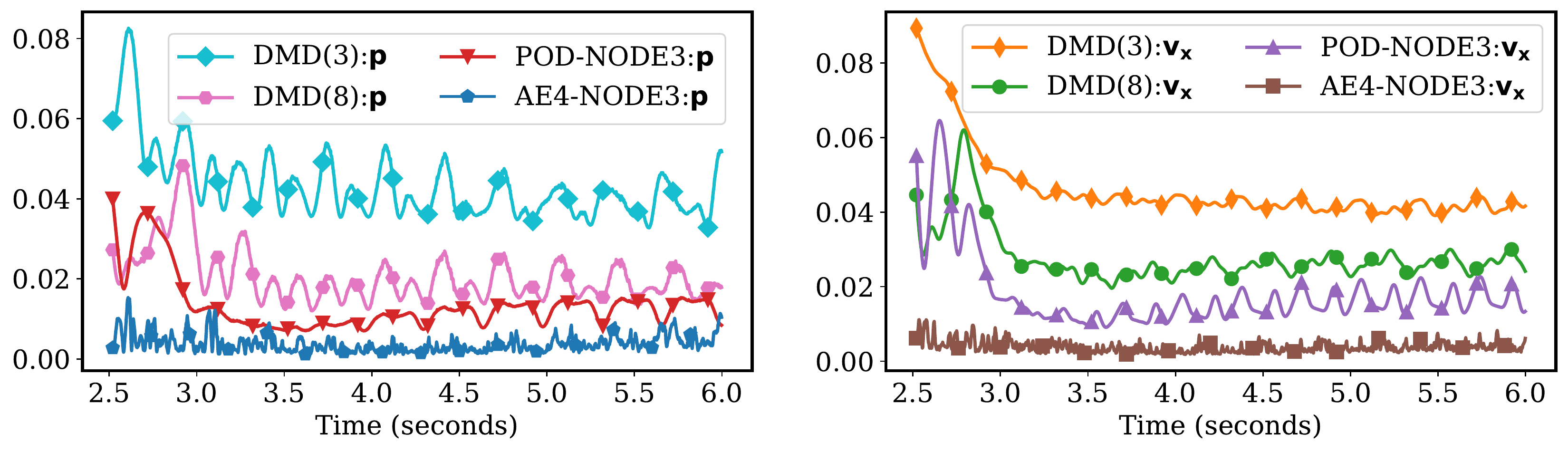}
    \end{center}
  \caption{Row1 - Comparison of RMSE of the NODE3 and the NODE5 models using POD and the AE1 basis; Row2 - Comparison of RMSE of the POD-NODE and the AE-NODE models with two DMD models with comparable latent space dimensions for the cylinder example}\label{fig:cylinder_aepod_dmd_rmse}
\end{figure}

The first row of Fig. \ref{fig:cylinder_aepod_dmd_rmse} compares the time trajectory of the spatial root mean square errors (RMSE) in the high-fidelity space for NIROM solutions generated using the POD and the AE1 basis with the NODE3 and NODE5 configurations. It is encouraging to note that even though the latent-space sizes are identical between the POD and the AE1 basis, the AE1-NODE solutions are  more accurate, confirming that the nonlinear AE basis generates a more accurate spatial compression than a linear POD basis of similar size. 
The second row of Fig. \ref{fig:cylinder_aepod_dmd_rmse} compares the RMSE between the AE4-NODE3, POD-NODE3, and two DMD NIROM models with truncation levels of $r=3$ and $r=8$. The AE4-NODE3 model achieves better accuracy than all the other models even while using the most compressed latent space ($p,v_x,v_y$:2,3,3). The DMD(8) model is roughly comparable with the POD-NODE3 model owing to the similarity in their latent space dimensions, whereas the DMD(3) model fares the worst due to the lack of hierarchical ordering in its latent basis.

\subsection{Shallow water equations}
The final numerical example involves flows governed by the depth-averaged SWE which is written in a conservative residual formulation as 
\begin{equation}\label{eq:conservative_compact}
\strongRes \equiv \pd{\vec q}{t} + \pd{\vec{p}_x}{x} + \pd{\vec{p}_y}{y} + \vec r = 0.
\end{equation}
Here, the state variable $\vec {q} = [h,u_x h,u_y h]^T$ consists of the flow depth, $h$, and the discharges in the $x$ and $y$ directions, given by $u_x h$ and $u_y h$, respectively. Further details about the flux vectors $\vec{p}_x$, $\vec{p}_y$ and the high-fidelity model equations are available in \cite{DFPSP2021}. The high-fidelity numerical solutions of the SWE are obtained using the 2D depth-averaged module of the Adaptive Hydraulics (AdH) finite element suite, which is a U.S. Army Corps of Engineers (USACE) high-fidelity, finite element resource for 2D and 3D dynamics \cite{Trahan2018}.

\subsubsection{Tidal flow in San Diego bay}
This numerical example involves the simulation of tide-driven flow in the San Diego Bay in California, USA.
\begin{wrapfigure}{R}{0.6\textwidth}
\centering
 \subfloat[POD-NODE $u_x$\label{fig:sd_rbf_u}]{%
   \includegraphics[width=0.28\columnwidth]{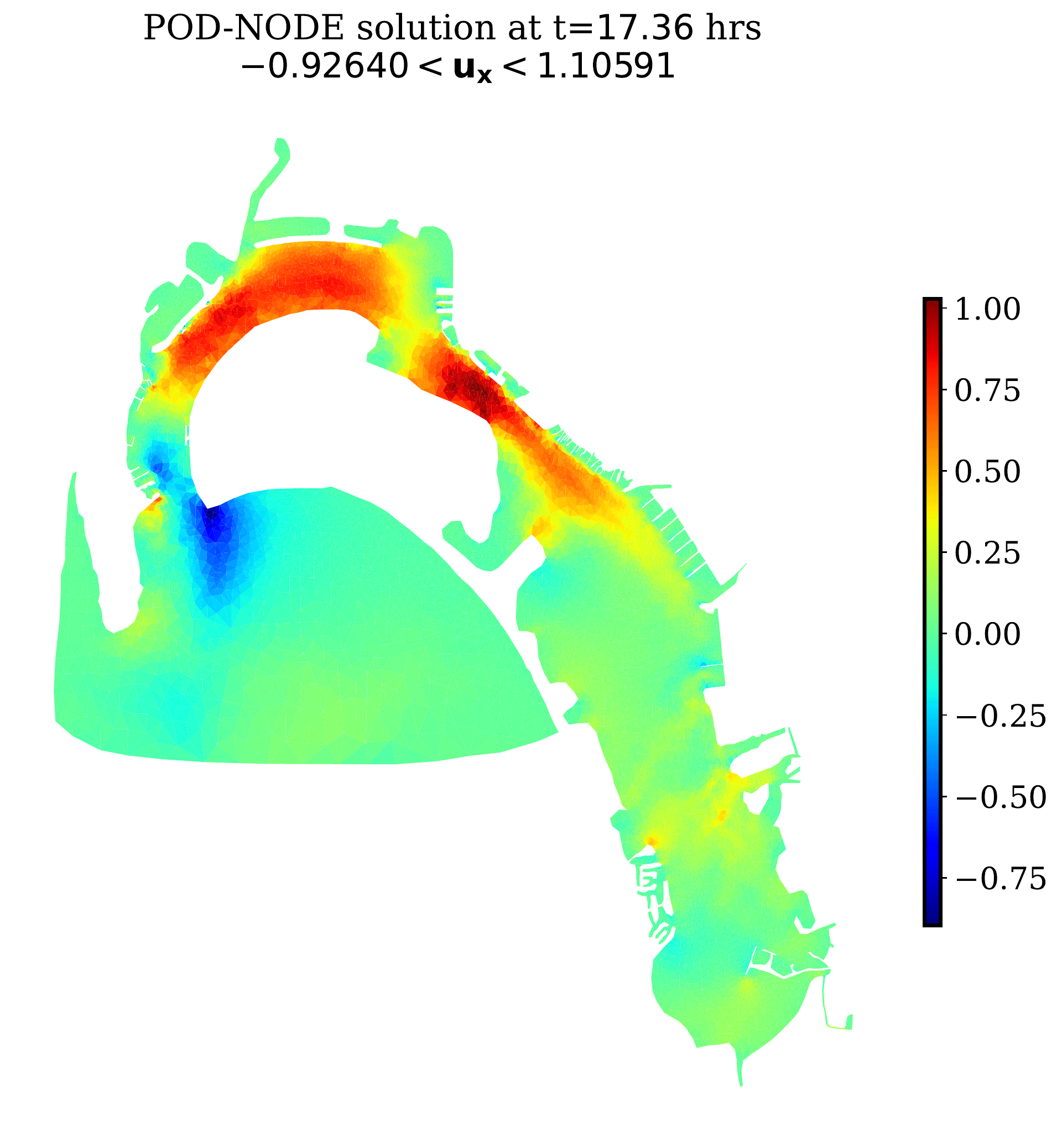}}
 \subfloat[AE-NODE $u_x$\label{fig:sd_node_u}]{%
   \includegraphics[width=0.28\columnwidth]{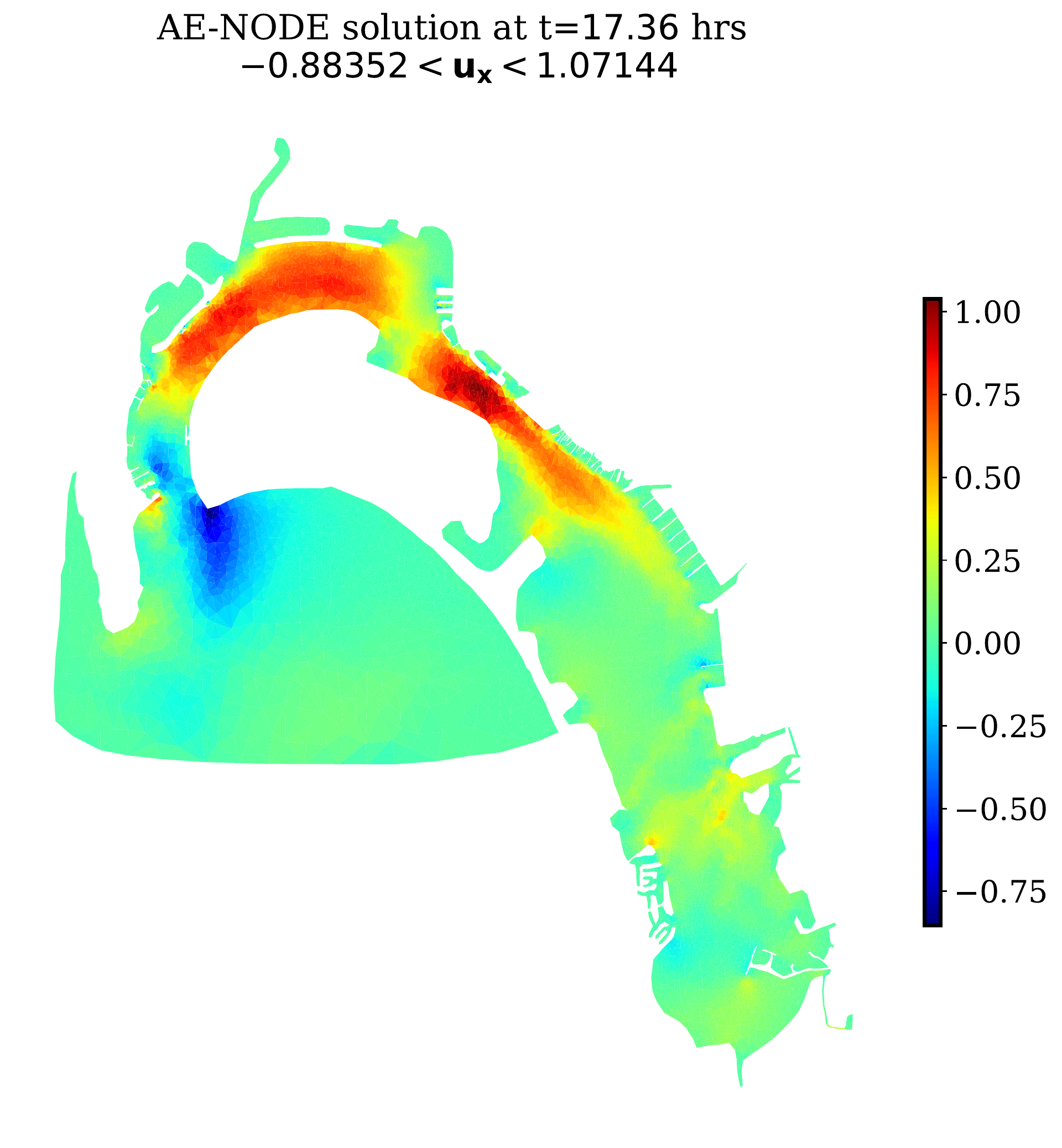}}\\
 \subfloat[POD-NODE error\label{fig:sd_rbf_uerr}]{%
   \includegraphics[width=0.28\columnwidth]{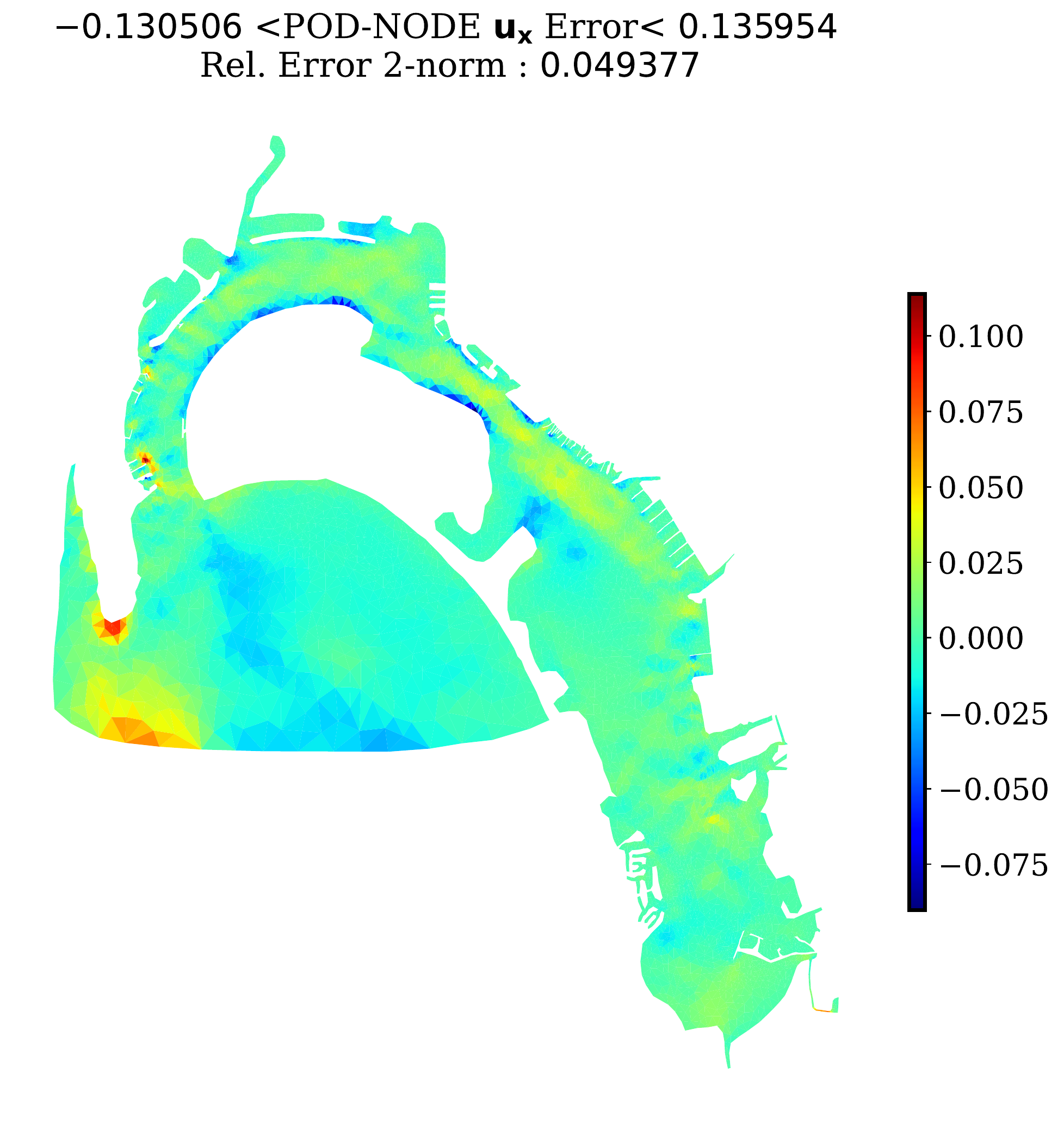}}
 \subfloat[AE-NODE error\label{fig:sd_node_uerr}]{%
   \includegraphics[width=0.28\columnwidth]{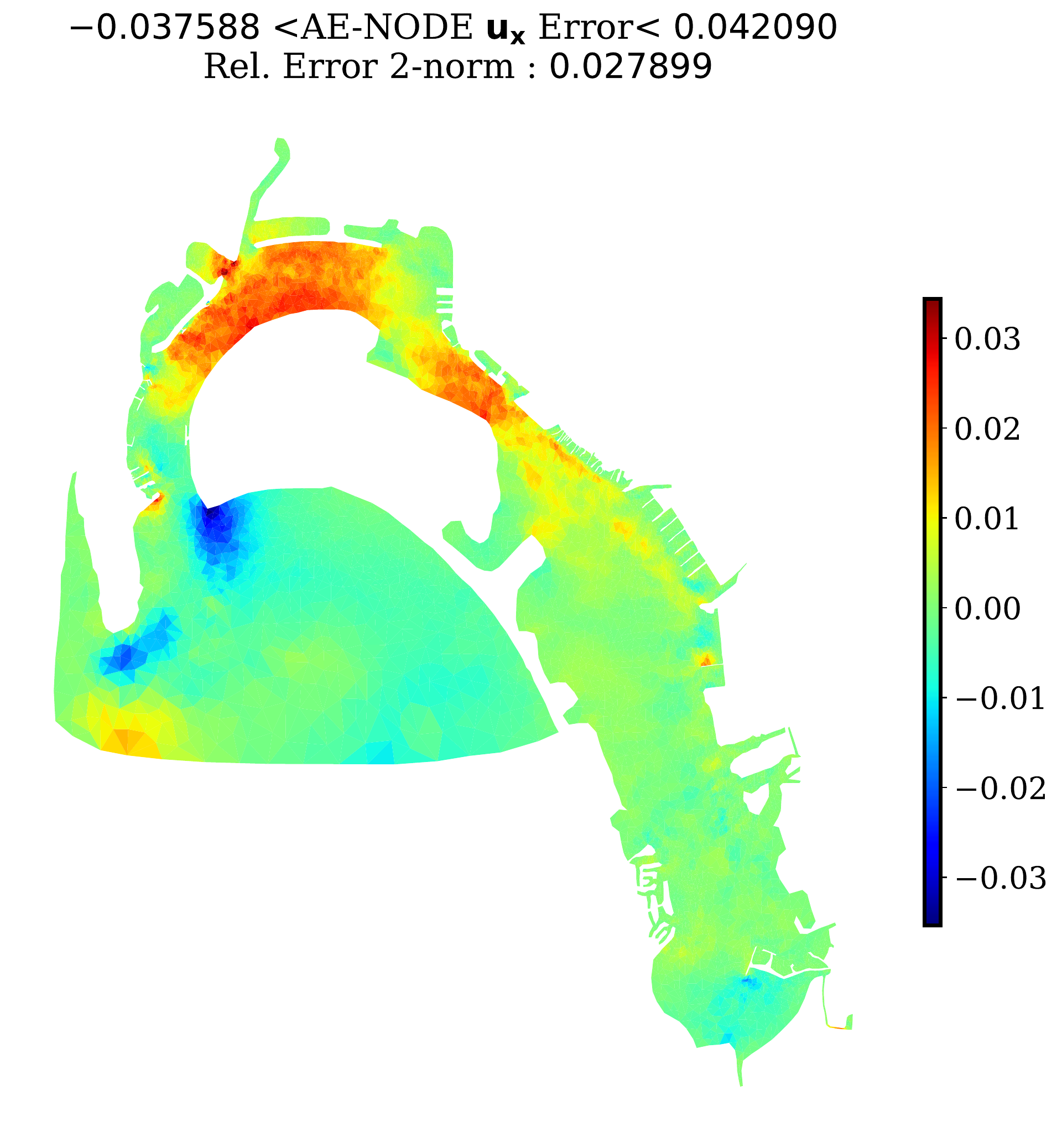}}
 \caption{NIROM solutions of $u_x$ and errors at $t=17.36$ hours for the San Diego example}\label{fig:sd_uplots}
\end{wrapfigure}
The AdH high-fidelity model consists of $N = 6311$ nodes, uses tidal data obtained from NOAA/NOS Co-Ops website at a tailwater elevation inflow boundary and has no flow boundary conditions everywhere else. Further details are available in \cite{DFPSP2021}.

The training space is generated using $1801$ high-fidelity snapshots obtained between $t=41$ minutes to $t=50$ hours at a time interval of $\Delta t=100$ seconds. The predicted ROM solutions are computed for the same time interval with $\Delta t=50$ seconds. A latent space of dimension $265 (p,u_x,u_y:36,115,113)$ is generated by using a POD truncation tolerance of $\tau_{POD}= 5\times10^{-7}$ for each solution component. The AE2 architecture designed for the cylinder example is modified by including \textit{BatchNormalization} layers for each hidden layer, using full batches for training, and by enforcing a latent space of dimension $30$ for each solution component. The RBF NIROM approximation is computed using a shape factor, $c = 0.01$. The simulation time points provided as input to the NODE model are normalized to lie in $t \in [0,1]$. The `dopri5' ODE solver is adopted for computing the hidden states both forward and backward in time. Learning from the conclusions of the cylinder example, a network consisting of a single hidden layer with $256$ neurons is deployed and the RMSProp optimizer with an initial learning rate of $0.001$, a staircase decay rate of $0.5$ every $5000$ epochs, and a momentum of $0.9$ is utilized for training the model over $20000$ epochs. For the DMD NIROM, the simulation time points are normalized to an unit time step, and a truncation level of $r=115$ is used to compute the DMD eigen-spectrum.

\begin{figure}[htb]
  \begin{center}
    \includegraphics[width=0.9\columnwidth]{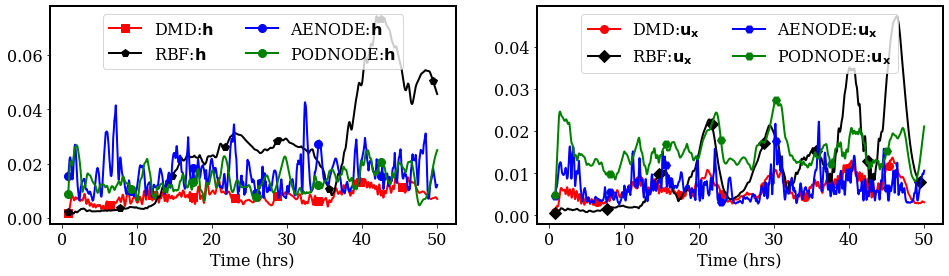}
  \end{center}
  \caption{NIROM RMSEs for the San Diego example}\label{san_diego_rms}
\end{figure}

Figure \ref{fig:sd_uplots} compares the POD-NODE and the AE-NODE NIROM solution fields (top row) for $u_x$ at $t=17.36$ hours and the corresponding error plots (bottom row). It can be seen that even while using a latent dimension that is three times smaller than POD, the relative errors for the AE-NODE solution ($0.0279$) is almost two times lower than that of the POD-NODE solution ($0.0494$).
Figure \ref{san_diego_rms} shows the spatial RMSE over time for the depth (left) and the x-velocity (right) NIROM solutions. The AE-NODE ($30$ modes) solution has comparable accuracy to the POD-NODE ($36$ modes) and the DMD ($115$ modes) solution for the depth variable and actually outperforms the POD-NODE ($115$ modes) solution for the x-velocity variable. Additionally, unlike the RBF NIROM solution, the AE-NODE solution does not exhibit any appreciable accumulation of error over time due to the higher-order time-stepping scheme adopted for NODE.

\section{CONCLUSION}
We have studied the combined use of autoencoders as a data-driven method for identifying an efficient reduced latent space that approximates the solution manifold of a system of nonlinear, time-dependent PDEs and neural ODEs as a non-intrusive method for modeling the evolution of the reduced latent space coefficients for the aforementioned dynamical system. Numerical experiments were carried out with a benchmark flow problem governed by the incompressible Navier Stokes equations and a real-world application of estuarine flow dynamics governed by the two-dimensional shallow water equations. The AE latent space representation was shown to provide a very high degree of efficient spatial compression, especially for the advection-dominated shallow water example. The POD-NODE NIROM formulation demonstrated a stable and accurate learning trajectory in modeling reduced basis dynamics, even in comparison to classical ROM techniques utilizing DMD, POD and RBF interpolation. 
The AE-NODE formulation also produced encouraging extrapolatory predictions for the flow around a cylinder example. This presents an exciting prospect for future exploration as even for an isolated system, unperturbed by unseen external forcings, truly extrapolative predictions of reduced order dynamics in flow regimes that do not correspond to the training data is a rare feature for most well-established ROM frameworks.

This study leads to several promising avenues of research. For instance, neural architecture search (NAS) tools can be adopted to perform an exhaustive exploration of the network architecture and model hyperparameter space for a wide range of flow dynamics in order to gain insight of the learning trajectory and to design more generalizable AE models with improved reconstruction accuracy. Moreover, integrating the process of identification of an AE-based latent space with the modeling of system dynamics using NODE may lead to significant performance improvements if the two independent learning problems can be designed to intelligently inform each other. 
All the relevant data and codes for this study will be made available in a public repository at \url{https://github.com/erdc/aenode_nirom} upon publication.

\section{ACKNOWLEDGMENTS}
This research was supported in part by an appointment of the first author to the Postgraduate Research Participation Program at the U.S. Army Engineer Research and Development Center, Coastal and Hydraulics Laboratory (ERDC-CHL) administered by the Oak Ridge Institute for Science and Education through an interagency agreement between the U.S. Department of Energy and ERDC. Permission was granted by the Chief of Engineers to publish this information.

\small
\bibliographystyle{hunsrt}
\bibliography{main.bib}

\end{document}

%% file: alt_Intro.tex
The computational challenges faced during \textit{high-fidelity} numerical simulations of engineering systems governed by nonlinear partial differential equations (PDEs), especially in applications involving control \cite{PBK2016}, optimal design and multi-fidelity optimization \cite{PWG2016}, can often be mitigated by the development of \textit{reduced order models} (ROMs)\cite{Benner_Gugercin_etal_15}.


\textit{Proper orthogonal decomposition} (POD) \cite{Sirovich_87,BHL1993} is a well known method for extracting a solution-dependent reduced basis space from a set of well-resolved, high-fidelity snapshots, and is most effective when the coherent structures of the dataset can be ranked in terms of their energy content. The POD method has been successfully applied in statistics \cite{J1986}, signal analysis and pattern recognition \cite{DM2008}, ocean models \cite{VH2006}, air pollution models \cite{FZPPBN2014}, convective Boussinesq flows \cite{SB2015}, and Shallow Water Equation (SWE) models {\cite{SSN2014,LFKG2016}}. Alternatively, nonlinear dimension reduction techniques such as kernel POD \cite{Salvador2021} or deep learning-based approaches like autoencoders \cite{Lusch2018, Ghorbanidehno2021} have also been used for extracting a reduced basis.
Combining autoencoder-generated bases with various specialized machine learning algorithms for time series modeling result in fully non-intrusive reduced order models \cite{Gonzalez_Balajewicz_2018, Eivazi2020,Maulik2021}. Hybrid methods \cite{Lee2020,Kim2020} can also be obtained by combining a nonlinear manifold learning technique like autoencoder for discovering the latent space with an intrusive method for the temporal dynamics.

Following the identification of the latent space, a reduced representation of the dynamical system is obtained by a Galerkin or Petrov-Galerkin projection on to the latent space \cite{LFKG2016,LFK2017}, which typically involves \textit{intrusive} modifications of the high-fidelity system operators. This work focuses on non-intrusive reduced order models (NIROMs) that do not require any knowledge of the high-fidelity simulator. In such a framework, the evolution of the expansion coefficients in the latent space is usually computed by the application of different regression-based methods directly on the high-fidelity data. These include artificial neural networks (ANNs), in particular multi-layer perceptrons \cite{HU2018}, Gaussian process regression (GPR) \cite{GH2019}, and radial basis function (RBF) \cite{ADN2013} interpolation. RBF interpolation in particular has been shown to be quite successful for nonlinear, time-dependent partial differential equations (PDEs) \cite{DFPSP2021}, nonlinear, parametrized PDEs \cite{ADN2013}, and aerodynamic shape optimization \cite{IQ2013}.

Alternatively, in deep neural networks (DNN) such as ResNet, the evolution of features over the depth of the network is equivalent to solving an ordinary differential equation (ODE) of the form $\frac{dz}{dt} = F(z,\theta)$ with the forward Euler method \cite{Ruthotto2019}.  With this connection in mind, \cite{Chen2018} proposed a 'continuous-depth' neural network called ODE-Net that effectively replaced the layers in ResNet-like architectures with a trainable ODE solver.  This neural ordinary differential equation approach (NODE) was further improved in \cite{Gholami2019,Dupont2019} and \cite{Finlay2020} proposed a NODE generative model that can be efficiently trained on large-scale datasets.  Some applications of the NODE framework include latent space closure modeling \cite{Maulik2020}, ODE/PDE model identification \cite{Sun2020}, modeling of irregularly spaced time series data \cite{RCD2019}, and modeling of spatio-temporal information in video signals \cite{KVKP2019}.  

Dynamic mode decomposition (DMD) is yet another method for obtaining a reduced order model. DMD represents the temporal dynamics of a complex, nonlinear system \cite{Schmid2010,KBBP2016} as the combination of a few linearly evolving, spatially coherent modes that oscillate at a fixed frequency, and which are closely related to the eigenvectors of the infinite-dimensional Koopman operator \cite{K1931}. Several variants of the DMD algorithm have been proposed \cite{PBK2016,KFB2016,ABBN2016,LV2017} and have been successfully applied as efficient ROM techniques for determining the optimal global basis modes for nonlinear, time-dependent problems \cite{DMD,Bistrian2017}. For non-parametrized PDEs, DMD presents an efficient framework that combines all the three stages of ROM development to learn a linear operator in an optimal least square sense. However, this approach cannot be directly applied to parametrized problems \cite{Alsayyari2021}.

In \cite{Dutta2021-AAAI}, we explored propagating the dynamics of a latent space formed from POD modes with neural ODEs. The present work investigates substituting the latent-space described by POD modes with one learned by an autoencoder. Our results for the combined autoencoder-NODE approach are compared to other methods like - a) dimension reduction by POD modes with latent space temporal dynamics captured by Neural ODEs (POD-NODE), b) dimension reduction via POD and temporal evolution of the latent space with Radial Basis Functions (POD-RBF), 
and c) Dynamic Mode Decomposition (DMD), which serve as benchmarks in our numerical experiments.  The performance of each approach will be evaluated on sample problems based on incompressible flow around a cylinder and shallow water hydrodynamics in the context of fast replay applications for complex fluid-dynamics problems.